\documentclass{article}

\usepackage[preprint]{neurips_2025}

\usepackage[utf8]{inputenc} % allow utf-8 input
\usepackage[T1]{fontenc}    % use 8-bit T1 fonts
\usepackage{hyperref}       % hyperlinks
\usepackage{url}            % simple URL typesetting
\usepackage{booktabs}       % professional-quality tables
\usepackage{amsfonts}       % blackboard math symbols
\usepackage{nicefrac}       % compact symbols for 1/2, etc.
\usepackage{microtype}      % microtypography
\usepackage{inconsolata}
\usepackage{caption}
\usepackage{amsmath}
\usepackage{amssymb}
\usepackage{mathtools}
\usepackage{enumitem}
\usepackage{makecell} 
\usepackage[usestackEOL]{stackengine}
\usepackage{graphicx}
\usepackage{capt-of}% or \usepackage{caption}
\usepackage{booktabs}
\usepackage{varwidth}
\usepackage[table,dvipsnames,svgnames]{xcolor}

\usepackage{tikz}
\usetikzlibrary{shadows}
\usepackage{float}
\usepackage{caption}
\usepackage{subcaption}
\usepackage{xspace}
\usepackage{svg}

\usepackage{import}
\usepackage{animate}
\usepackage{arydshln}
\usepackage{multirow}
\usepackage[normalem]{ulem}
\usepackage[most]{tcolorbox}
\usepackage{colortbl}
\usepackage{tcolorbox}
\usepackage{pifont}
\usepackage{alltt}
\definecolor{titlegray}{rgb}{0.4, 0.4, 0.4} % A medium-dark gray for title/frame
\definecolor{contentgray}{rgb}{0.95, 0.95, 0.95} % A very light gray for content background
\usepackage{algorithm}
\usepackage{algpseudocode}

\newcommand{\model}{\text{Video-MOPD-8B}\xspace}
\newcommand{\modelbf}{\textbf{Video-MOPD-8B}\xspace}

\definecolor{carolinablue}{rgb}{0.6, 0.73, 0.89}
\definecolor{mildgreen}{rgb}{0.85, 0.98, 0.80}
\definecolor{beautycolor}{rgb}{0.91, 0.75, 0.96} % darker purple
\definecolor{fallacycolor}{rgb}{0.85, 0.95, 1}
\definecolor{gendercolor}{rgb}{1, 0.85, 0.85}
\definecolor{brightyellow}{RGB}{255, 255, 100}
\definecolor{boxcolor}{RGB}{51,51,153}
\definecolor{lightgreen}{rgb}{0.56, 0.93, 0.56}
\definecolor{citeblue}{HTML}{0064E0}

\hypersetup{
    colorlinks=true,
    citecolor=citeblue,
    linkcolor=red, % 可选
    urlcolor=citeblue % 可选
}

\definecolor{deepblue}{RGB}{0, 0, 139}

\title{Video-MOPD: Multi-Teacher On-Policy Distillation for Video Understanding}

\newtcolorbox{questionbanner}{
  colback=blue!10!white,    % background color
  colframe=blue!80!black,   % border color
  width=\textwidth,
  arc=4mm,                  % rounded corners
  boxrule=1pt,              % border thickness
  fonttitle=\bfseries,
  title=Question,
}
\newtcolorbox{promptbox}[1]{
  colback=contentgray,      % Background color of the content area
  colframe=titlegray,       % Color of the box frame
  colbacktitle=titlegray,   % Background color of the title bar (used when title is set)
  coltitle=white,           % Color of the title text (used when title is set)
  title={#1}, % <-- REMOVE or COMMENT OUT THIS LINE
  arc=4mm,                  % Radius of the rounded corners
  rounded corners=northwest, % Top-left rounded
  rounded corners=northeast, % Top-right rounded
  sharp corners=south,      % Bottom corners sharp
  boxrule=1pt,              % Thickness of the frame
  fonttitle=\bfseries,      % Make the title text bold
}

\definecolor{questionbg}{RGB}{240, 248, 255}  % 淡蓝色背景
\definecolor{answerbg}{RGB}{245, 255, 250}   % 淡绿色背景
\definecolor{bordercolor}{RGB}{100, 149, 237} % 边框颜色
\definecolor{titlecolor}{RGB}{25, 25, 112}    % 标题颜色

\newtcolorbox{vqaexample}[2][]{
    enhanced,
    breakable,
    colback=white,
    colframe=bordercolor,
    boxrule=1.5pt,
    arc=4pt,
    outer arc=4pt,
    left=8pt,
    right=8pt,
    top=8pt,
    bottom=8pt,
    drop shadow={shadow xshift=0.5mm, shadow yshift=-0.5mm, opacity=0.3},
    overlay={
        \node[
            anchor=north east,
            xshift=-3pt,
            yshift=-3pt,
            fill=bordercolor!80,
            text=white,
            font=\bfseries,
            rounded corners=2pt,
            inner sep=4pt,
            minimum height=1.2em,
            align=center
        ] at (frame.north east) {#2};
    },
    #1
}

\author{
Zhenxin Qin$^{1,2,*}$,\hspace{0.45em}
Peng Shi$^{*,\dagger}$,\hspace{0.45em}
Cong Han$^{2}$,\hspace{0.45em}
Yinlong Qian$^{2}$,\hspace{0.45em}
Zequn Jie$^{2,\ddagger}$,\hspace{0.45em}
Lin Ma
\\[8pt]
$^1$\textbf{Tongji University}\hspace{2em}$^2$\textbf{Bilibili Inc.}
}
\begin{document}
\raggedbottom

\maketitle
\renewcommand{\thefootnote}{}
\footnotetext{$^*$Equal contribution, $^\dagger$Project leader, $^\ddagger$Corresponding author.}
\vspace{-8mm}
{\centering\small \{zhenxinqin1, pengshi.scholar, hancong0911, zequn.nus, forest.linma\}@gmail.com\par}
\vspace{6mm}

% !TEX root = ../neurips_2025.tex
\begin{abstract}
Video understanding demands a convergence of complementary capabilities across perception, temporal understanding, and complex reasoning, which are difficult to jointly optimize within a single model. We introduce \modelbf, an open-weight model dedicated to video understanding tasks. To fundamentally enhance its capabilities, we conduct targeted reinforcement learning (RL) optimization across three core domains: video temporal grounding (VTG), general video comprehension, and video STEM reasoning. We then unify their complementary capabilities via Multi-Teacher On-Policy Distillation (MOPD), which consolidates expert knowledge by supervising student-generated trajectories with routed teacher feedback. We further introduce \textbf{Reliability-Aware Informative Sampling (RAIS)}, which selects examples with consistently reliable teacher supervision and large teacher--student performance gaps. Together, these components enable \model to achieve coordinated and comprehensive performance gains across diverse video understanding tasks. Extensive experiments on comprehensive benchmarks covering general video understanding, temporal grounding, video reasoning, and video STEM tasks demonstrate that \model achieves state-of-the-art performance among existing models at a comparable scale. The trained model weights are available at \url{https://huggingface.co/LandH/Video-MOPD-8B}.
\end{abstract}

\section{Introduction}
\label{sec:introduction}

Recent advances in multimodal large language models (MLLMs) have substantially expanded visual understanding, enabling increasingly capable perception and reasoning over both images and videos~\citep{li2024llavaonevisioneasyvisualtask,bai2025qwen3vltechnicalreport,li2026videochat3fullyopenvideo}. Compared with static images, videos introduce continuously evolving visual content, temporal dependencies, and complex interactions among objects and events. As video MLLMs continue to improve through stronger visual encoders, large-scale multimodal instruction tuning, and more effective post-training, the focus of video understanding is moving beyond basic question answering toward broader and more fine-grained capabilities~\citep{Pan_2026_CVPR,fu2026videommev2stagebenchmarkscomprehensive}. A comprehensive video model is expected not only to understand overall video semantics, but also to accurately associate language with temporal events and reason over specialized or knowledge-intensive visual content.

In practice, these requirements involve several complementary aspects of video intelligence. \emph{General video understanding} requires models to recognize actions, events, object interactions, temporal relations, and causal dependencies across diverse scenarios. \emph{Video temporal grounding} (VTG) places greater emphasis on fine-grained temporal perception by requiring the model to identify the precise segment corresponding to a natural-language query~\citep{zhu2026timelens2generalistvideotemporal}. Beyond general-purpose perception and localization, \emph{video STEM reasoning} requires the model to integrate visual-temporal evidence with scientific or technical knowledge and perform multi-step reasoning~\citep{hu-etal-2026-video,scivideobench}. These capabilities emphasize different aspects of video understanding, ranging from broad semantic comprehension and precise temporal alignment to knowledge-intensive reasoning. Consequently, a general-purpose video model needs post-training strategies that can effectively strengthen these complementary capabilities.

Alongside advances in architecture and multimodal training data, reinforcement learning (RL) has emerged as an effective post-training paradigm for improving specific multimodal capabilities~\citep{Yang_2026_CVPR,liu2026lengthunbiasedsequencepolicyoptimization,Zhang_2026_CVPR}. Recent work demonstrates that reinforcement fine-tuning can substantially improve video perception and reasoning~\citep{qin2026easyvideor1easierrlvideo,Wang_2026_CVPR,liu2026reasoningintersectionconsensusframealignment}. In particular, task-specific rewards make it possible to directly optimize capabilities that are difficult to capture through supervised fine-tuning alone~\citep{luo-etal-2026-museg}. Motivated by these advances, we perform targeted optimization to strengthen three complementary aspects of video understanding: \textbf{general video understanding}, \textbf{video temporal grounding}, and \textbf{video STEM reasoning}. We construct dedicated training data and optimization objectives for each capability, producing specialized models with complementary strengths.

Although domain-specific optimization can produce strong specialist models, it naturally results in multiple specialized models. Deploying separate models for different tasks is undesirable: a general-purpose video model should retain their complementary improvements within a single set of parameters. To achieve this goal, we adopt \emph{Multi-Teacher On-Policy Distillation} (MOPD)~\citep{ma2026mopdmultiteacheronpolicydistillation}. MOPD distills multiple domain-specialized teachers into a shared student using trajectories sampled from the student policy itself, with the corresponding teacher providing dense token-level supervision. Recent video-specific work applies on-policy distillation to temporal video grounding~\citep{li2026videoopd}, whereas our setting consolidates heterogeneous capabilities from multiple domain-specialized teachers into a single video model. Unlike off-policy distillation based on teacher-generated trajectories, MOPD provides teacher supervision on states visited by the student policy itself. This property makes MOPD particularly suitable for consolidating capabilities that have been independently strengthened through targeted post-training.

Based on this strategy, we present \modelbf, an open-weight model for comprehensive video understanding. Our specialize-then-unify pipeline consists of two stages. First, starting from a shared base model, we construct three complementary experts: a \emph{General Video Expert} for broad video comprehension and cross-frame reasoning, a \emph{VTG Expert} for fine-grained event localization and precise language--video temporal alignment, and a \emph{STEM Expert} for knowledge-intensive reasoning involving fine-grained perception, OCR, spatial understanding, mathematical reasoning, and scientific or technical knowledge. Second, we employ MOPD to distill these complementary capabilities into a shared student. Each training sample is routed to its matched domain teacher, which provides dense token-level supervision on the student-generated trajectory during on-policy training. To improve capability transfer, we further introduce \emph{Reliability-Aware Informative Sampling} (RAIS), which retains samples with consistently reliable teacher supervision and prioritizes those with larger teacher--student performance gaps. The resulting \modelbf integrates the improvements of multiple specialists while maintaining a unified architecture and inference interface. Figure~\ref{fig:mopd_framework} provides an overview of the complete framework.

We extensively evaluate \model on benchmarks covering general video understanding, temporal grounding, video reasoning, and video STEM tasks. Targeted optimization consistently improves the corresponding specialist capabilities, while MOPD consolidates their complementary strengths into a single model. Across seven benchmarks, \model improves the average score by 5.4 points over Qwen3-VL-8B-Instruct and reaches 69.12 overall, outperforming parameter averaging by 1.84 points while leading or tying six of seven benchmarks among non-expert baselines. These results demonstrate the effectiveness of combining domain specialization with multi-teacher on-policy distillation as a scalable post-training recipe for comprehensive video understanding. To facilitate further research and development, we release the trained model weights.

Our main contributions are summarized as follows:

\begin{itemize}[leftmargin=1.5em,labelsep=0.5em]

\item We present \modelbf, an open-weight model for comprehensive video understanding, built through a specialize-then-unify post-training pipeline that covers general video understanding, video temporal grounding, and video STEM reasoning.

\item We construct three complementary domain experts and consolidate their capabilities through domain-routed Multi-Teacher On-Policy Distillation. We further introduce Reliability-Aware Informative Sampling (RAIS), which filters for reliable teacher supervision and prioritizes samples with larger teacher--student performance gaps.

\item We conduct extensive evaluations across general video understanding, temporal grounding, video reasoning, and video STEM benchmarks. \model achieves state-of-the-art performance among existing models at a comparable scale, and we release its trained model weights.

\end{itemize}

\section{Related Work}
\label{sec:related_work}

\paragraph{Video Multimodal Large Language Models.}
Recent advances in multimodal large language models have substantially improved video understanding by extending image-based vision--language models to temporally structured visual inputs. Representative models such as LLaVA-OneVision~\citep{li2024llavaonevisioneasyvisualtask}, VideoLLaMA3~\citep{zhang2025videollama3frontiermultimodal}, InternVideo2.5~\citep{wang2025internvideo25empoweringvideomllms}, and Apollo~\citep{11094526} explore different aspects of video modeling, including unified image--video representation, temporal encoding, fine-grained visual perception, video sampling, and large-scale multimodal instruction tuning. Alongside model development, recent benchmarks have expanded video evaluation beyond general-purpose question answering toward more specialized and knowledge-intensive settings. MMVU~\citep{mmvu} evaluates expert-level video understanding across multiple academic disciplines, while Video-MMMU~\citep{hu-etal-2026-video} focuses on knowledge acquisition and application from professional videos. These developments highlight the increasingly diverse capabilities required for comprehensive video understanding, spanning general semantic comprehension, fine-grained temporal localization, and knowledge-intensive reasoning. This motivates targeted post-training strategies that can effectively strengthen these complementary capabilities while ultimately integrating them into a unified model.

\paragraph{Reinforcement Learning for Video Understanding.}
Reinforcement learning has recently emerged as an effective post-training approach for improving reasoning and task-specific capabilities in multimodal models. Video-R1~\citep{NEURIPS2025_8eb39768} extends R1-style reinforcement learning to video reasoning and demonstrates that reward-driven optimization can improve reasoning over dynamic visual content. VideoChat-R1~\citep{videochat_r1} further investigates reinforcement fine-tuning for spatiotemporal perception with task-specific rewards for video question answering, temporal grounding, and tracking. More specialized studies focus on fine-grained temporal understanding: MUSEG~\citep{luo-etal-2026-museg} improves temporal reasoning through timestamp-aware multi-segment grounding and customized reinforcement learning, while Tempo-R0~\citep{yue2025tempor0videomllmtemporalvideo} applies reinforcement learning to temporal video grounding with explicit temporal sensing. Beyond individual tasks, OneThinker~\citep{Feng_2026_CVPR} explores unified reinforcement learning across heterogeneous image and video tasks, including question answering, captioning, grounding, tracking, and segmentation. These works establish reinforcement learning as a flexible mechanism for enhancing video-specific capabilities. Video-MOPD organizes these advances into a specialize-then-unify pipeline: domain-specific RL first maximizes complementary capabilities, followed by multi-teacher consolidation into a unified model.

\paragraph{Capability Integration and On-Policy Distillation.}
Existing approaches to capability integration include mixed RL over multiple domains~\citep{yang2025qwen3technicalreport}, sequential domain-wise RL~\citep{wang2026nemotroncascade}, offline imitation from specialized teachers~\citep{liu2025deepseekv32}, and parameter-space model merging~\citep{model_soups,task_arithmetic}. These paradigms make different trade-offs: mixed and sequential RL couple capability integration with optimization across domains, offline imitation relies on teacher-generated trajectories that differ from the student's inference-time distribution, and parameter-space merging can encounter interference among task-specific parameter updates~\citep{ranzato2016sequence,yadav2023ties}. Multi-Teacher On-Policy Distillation (MOPD)~\citep{ma2026mopdmultiteacheronpolicydistillation} instead uses domain-specialized teachers to supervise trajectories generated by the student itself, enabling capability integration directly in policy space. In Video-MOPD, we adopt MOPD to consolidate the complementary capabilities of the General Video Expert, VTG Expert, and STEM Expert into the unified \modelbf.

\section{Method}
\label{sec:method}

\begin{figure}[H]
    \centering
    \includegraphics[width=\textwidth]{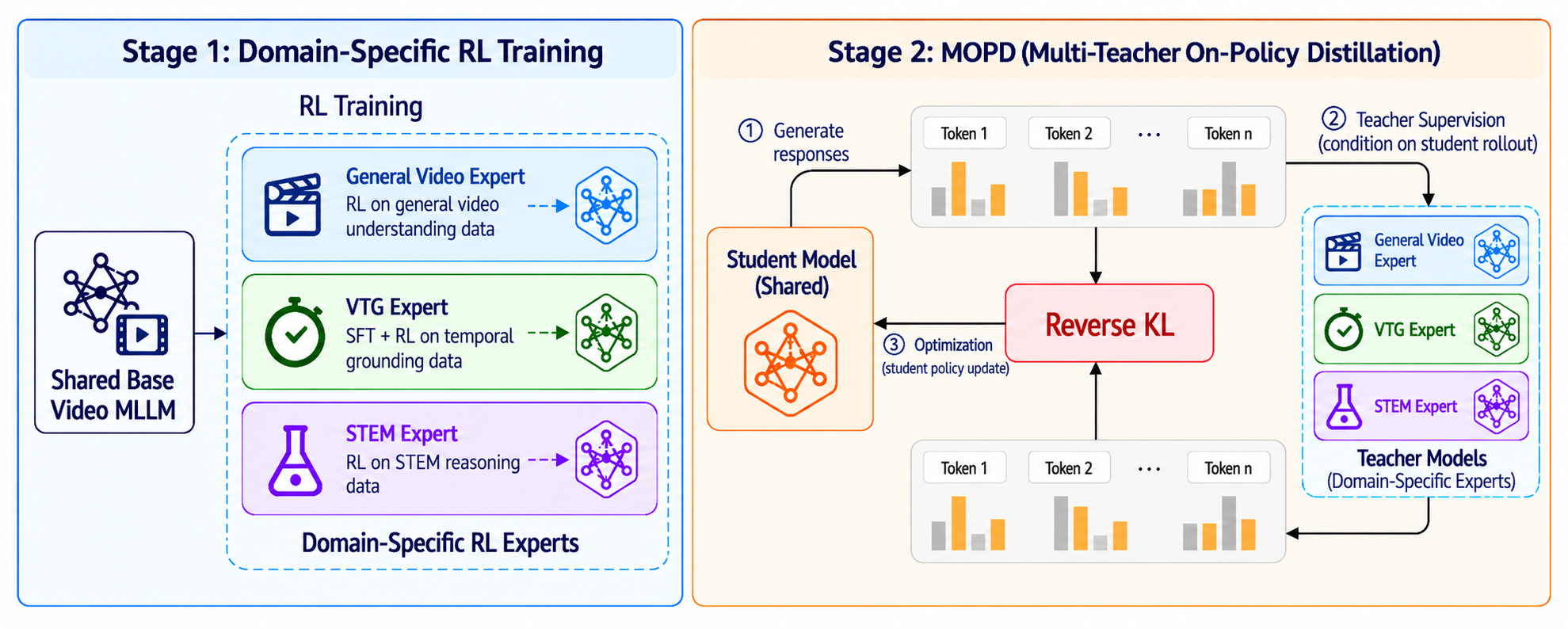}
    \caption{Overview of Video-MOPD. Three independently trained experts provide complementary capabilities. Each domain-labeled sample is routed to exactly one matched expert, which provides token-level supervision for the student's on-policy completion.}
    \label{fig:mopd_framework}
\end{figure}

Figure~\ref{fig:mopd_framework} illustrates the overall post-training pipeline of Video-MOPD. Starting from a shared base model, we first construct three complementary specialists: the \emph{General Video Expert}, \emph{VTG Expert}, and \emph{STEM Expert}. The General Video Expert and STEM Expert are optimized through domain-specific reinforcement learning, while the VTG Expert follows the TimeLens2 training recipe~\citep{zhu2026timelens2generalistvideotemporal}, adopting reinforcement learning with verifiable rewards for fine-grained temporal grounding. This stage produces three specialized teachers with complementary capabilities. We then employ Multi-Teacher On-Policy Distillation (MOPD)~\citep{ma2026mopdmultiteacheronpolicydistillation} to consolidate their capabilities into a single unified model. During distillation, each training sample is routed to its corresponding teacher, which provides dense token-level supervision over trajectories generated by the student policy. Through this specialize-then-unify pipeline, the resulting \modelbf integrates the gains obtained from multiple targeted post-training processes while retaining a unified architecture and inference interface.

\subsection{Construction of Specialized Teacher Experts}
\label{section:experts}

To build a comprehensive video-understanding model, we train three domain-specific experts, each specialized in a unique facet of video comprehension. Our design stems from a core observation: diverse video-understanding tasks call for divergent reasoning capacities and often demand distinct training paradigms. For example, STEM-related video reasoning relies on explicit thinking traces to decompose intricate problems, where chain-of-thought-aware training proves critical. By contrast, video temporal grounding targets precise event localization via reinforcement learning; it does not require explicit intermediate reasoning in its output and directly optimizes timestamp boundary prediction performance. Training independent experts with task-specific pipelines over carefully curated datasets allows each model to attain strong domain specialization. Meanwhile, this setup avoids objective conflicts that would occur when a single model is asked to fit mutually incompatible training requirements. In the subsequent subsections, we elaborate on the training data, optimization objectives, and task-specialized strategies for each expert. All three specialists are training-time teachers; deployment uses only the unified Video-MOPD student, with no additional expert or routing overhead.

\subsubsection{General Video Understanding Expert}
The general video understanding expert targets holistic semantic comprehension of video content, spanning visual entity recognition, spatial-temporal reasoning, and contextual event interpretation across diverse open-domain scenarios. We assemble a collection of public open-source video-understanding datasets spanning causal and situated reasoning, long-video understanding, large-scale instruction diversity, and diagnostic perception, including NeXT-QA~\cite{xiao2021nextqa}, LongVideo-Reason~\cite{chen2025longvilarn1}, STAR~\cite{wu2021star}, LLaVA-Video-178K~\cite{zhang2024llavavideo}, Holmes-train~\cite{cheng2025videoholmes}, PerceptionTest~\cite{patraucean2023perceptiontest}, CLEVRER~\cite{yi2020clevrer}, and SR-91k~\cite{ouyang2025spacer}. We filter out overly simple and excessively hard samples, empirically reweight each source according to its contribution to distinct capability dimensions, and apply benchmark-aware decontamination to avoid test-set leakage.

Initialized from Qwen3-VL-8B-Instruct, we optimize the expert via GRPO with on-the-fly difficulty filtering. The composite reward combines an answer-accuracy reward with a response-format compliance reward, jointly driving the model to produce verifiable reasoning traces alongside final predictions. This optimization strengthens short-video comprehension, particularly action recognition, event interpretation, and cross-frame visual-evidence aggregation.

\subsubsection{Video Temporal Grounding Expert}
Initialized from Qwen3-VL-8B-Instruct, our video temporal grounding (VTG) expert fully reproduces the two-stage training pipeline from TimeLens2~\cite{zhu2026timelens2generalistvideotemporal}. In the supervised fine-tuning (SFT) phase, we train on the combined dataset of TimeLens2-93K, TimeLens-100K~\cite{timelens}, and Ego4D-NLQ~\cite{ego4d}. Following the original setup, diverse instruction prompts and timestamp output formats are applied, so the model learns variable-cardinality interval-set prediction and disentangles genuine temporal localization ability from superficial prompt-format imitation.

We further boost localization performance via GRPO reinforcement learning equipped with rollout-guided hard-sample mining. The composite reward combines set-level temporal IoU and a temporal Wasserstein reward $R_{TW}$, which offers geometry-aware feedback for near-miss predictions even under zero-overlap conditions, alongside a penalty for unparseable outputs. The resulting expert attains strong multi-span fine-grained grounding ability, providing exclusive language-video temporal-alignment supervision for MOPD distillation.

\subsubsection{Video STEM Reasoning Expert}
Public post-training video datasets are largely short of discipline-oriented STEM video samples. Inspired by the insight that text-based STEM data can boost image-level STEM reasoning performance~\cite{qiu2025metisrise}, we leverage image-form STEM resources to facilitate discipline-oriented understanding for video inputs. We utilize the Orsta47K~\cite{ma2025one} and virl39K~\cite{wang2025vlrethinker} datasets adopted in METIS-SPECS~\cite{chen2025metisspecs}. We conduct extensive data processing operations on the raw corpora and construct a multi-ability dataset that covers STEM reasoning alongside other diverse visual capabilities.

Initialized from Qwen3-VL-8B-Instruct, we adopt the DAPO variant of GRPO~\cite{yu2025dapo}, building on the GRPO objective introduced by DeepSeekMath~\cite{shao2024deepseekmath}, for model optimization. Although the whole training is conducted purely on static image examples, the expert acquires critical frame-level competencies, including OCR, chart parsing, geometric analysis, and multi-step mathematical reasoning, that transfer naturally to individual video frames and establish essential prerequisites for video-based STEM tasks.

\subsection{Reliability-Aware Informative Sampling (RAIS)}
\label{sec:data_selection}

The MOPD training set consists of Video, Image, and temporal-grounding examples, which are routed to the General Video Expert, STEM Expert, and VTG Expert, respectively. We use approximately balanced domain-level sampling to reduce imbalance caused purely by differences in dataset size. Each example $x$ is associated with a domain label $d(x)$, which determines the corresponding teacher during MOPD training.

Naive random sampling may retain examples on which the corresponding teacher is unreliable or the initial student already performs well, resulting in unreliable supervision or limited learning value. We therefore adopt RAIS, which jointly considers teacher consistency and the remaining difficulty for the student. For each candidate example $x$, we independently sample $K$ responses from the corresponding domain teacher $\pi_T^{d(x)}$ and the initial student $\pi_S$, denoted by $\{y_k^T\}_{k=1}^{K}$ and $\{y_k^S\}_{k=1}^{K}$, respectively. Let $V_d(x,y)\in\{0,1\}$ denote the task-specific correctness verifier for domain $d$. We estimate the empirical success rates of the teacher and student as
\begin{equation}
    \mathrm{Acc}_T(x)
    =
    \frac{1}{K}
    \sum_{k=1}^{K}
    V_{d(x)}(x,y_k^T),
    \qquad
    \mathrm{Acc}_S(x)
    =
    \frac{1}{K}
    \sum_{k=1}^{K}
    V_{d(x)}(x,y_k^S).
\end{equation}

We first apply a reliability filter and retain only examples with $\mathrm{Acc}_T(x)=1$, requiring the corresponding teacher to solve the example correctly across all $K$ sampled responses. This reliability criterion ensures that every retained example is supported by consistent teacher supervision across all $K$ rollouts.

For the retained examples, we further quantify their remaining learning value using the teacher--student performance gap
\begin{equation}
    \Delta(x)
    =
    \mathrm{Acc}_T(x)-\mathrm{Acc}_S(x)
    =
    1-\mathrm{Acc}_S(x).
\end{equation}
We prioritize examples with larger $\Delta(x)$, which exhibit a larger unresolved teacher--student performance gap despite consistent teacher success. In this way, the selected MOPD training data simultaneously provide reliable teacher supervision and informative learning signals for effective capability transfer.

\subsection{Video-MOPD: Routed Multi-Teacher On-Policy Distillation}
\label{sec:mopd_objective}

Let $\mathcal{D}=\bigcup_{d\in\{v,i,t\}}\mathcal{D}_d$ denote the mixed training set consisting of video, image, and temporal-grounding examples. The three subsets correspond to the General Video Expert, STEM Expert, and VTG Expert, respectively. Each domain $d$ is associated with a specialized teacher policy $\pi_T^d$, while $\pi_\theta$ denotes the unified student. For each training sample $x$, a deterministic router $r(x)\in\{v,i,t\}$, specified by its domain label, selects the corresponding teacher $\pi_T^{r(x)}$. The teachers and router are used only during training and are removed after distillation.

For an input $x\sim\mathcal{D}$, the student generates an on-policy trajectory
\begin{equation}
    y=(y_1,\ldots,y_{|y|})
    \sim \pi_\theta(\cdot\mid x).
    \label{eq:on_policy_rollout}
\end{equation}
The generated trajectory is then routed to the matched domain teacher, which evaluates the same token sequence under the same context. Following MOPD~\citep{ma2026mopdmultiteacheronpolicydistillation}, we optimize the student toward the routed teacher using a token-level reverse-KL objective:
\begin{equation}
\begin{aligned}
    \mathcal{L}_{\mathrm{Video\text{-}MOPD}}(\theta)
    &=
    \sum_{d\in\{v,i,t\}} p_d\,
    \mathbb{E}_{\substack{x\sim\mathcal{D}_d,\\
    y\sim\pi_\theta(\cdot\mid x)}}
    \left[
        \frac{1}{|y|}
        \sum_{t=1}^{|y|}
        D_{\mathrm{KL}}
        \left(
            \pi_\theta(\cdot\mid x,y_{<t})
            \,\|\,
            \pi_T^d(\cdot\mid x,y_{<t})
        \right)
    \right].
\end{aligned}
    \label{eq:video_mopd}
\end{equation}
where $p_d$ denotes the sampling proportion of domain $d$, and $|y|$ denotes the number of tokens in the student-generated response. The reverse-KL objective provides dense token-level supervision on trajectories generated by the student itself, allowing each domain expert to guide the student only on samples matched to its specialization.

All three domains jointly update the same student throughout training. In this way, Video-MOPD consolidates the complementary capabilities of the General Video Expert, STEM Expert, and VTG Expert into a single unified model rather than combining their predictions at inference time. The resulting \modelbf performs inference independently with a single set of parameters.

\section{Experiments}
\label{section:experiment}

\subsection{Experimental Setup}

\paragraph{Benchmarks.}
We evaluate Video-MOPD across seven complementary video benchmarks spanning STEM reasoning, general video understanding, and temporal understanding. \textbf{MVBench}~\citep{Li_2024_CVPR} contains 4,000 multiple-choice questions for general video understanding. \textbf{MMVU}~\citep{mmvu} evaluates knowledge-intensive and open-ended multimodal reasoning; we report the GPT-5-judged overall score. \textbf{Video-MME}~\citep{11093290} measures broad video perception and reasoning. \textbf{VideoMMMU}~\citep{hu-etal-2026-video} targets discipline-level video understanding, for which we report GPT-5 judge accuracy. \textbf{Video-Holmes}~\citep{cheng2025videoholmes} stresses complex video reasoning and evidence integration. \textbf{TimeLens}~\citep{timelens} reports the arithmetic mean of temporal intersection-over-union on Charades, ActivityNet, and QVHighlights, while \textbf{TempCompass}~\citep{liu-etal-2024-tempcompass} evaluates fine-grained temporal perception. We preserve each benchmark's native evaluation protocol and report all seven scores individually; TimeLens uses temporal IoU, while the remaining benchmarks follow their standard scoring procedures.

\paragraph{Compared models.}
Our primary baseline is the common Qwen3-VL-8B-Instruct initialization. We compare \model with Qwen3-VL-8B-Thinking~\citep{bai2025qwen3vltechnicalreport}, CRPO~\citep{du2026learningspatiotemporalsensitivityvideo}, VideoKR~\citep{fu2026videokrknowledgereasoningintensivevideo}, OneThinker~\citep{Feng_2026_CVPR}, VideoSSR~\citep{He_2026_CVPR}, and each of the three domain experts. We additionally include \textbf{Param-Merge (Avg.)}, which directly averages the parameters of the three specialized expert checkpoints, as a parameter-space capability-integration baseline. The expert and model-merging comparisons evaluate capability consolidation rather than parameter scaling, since all models operate at the same nominal scale. For ablation, we compare sampled-token feedback with Teacher Top-16, Student Top-16, and Teacher--Student overlap Top-16 supports under matched one-epoch training.

\begin{figure}[H]
    \centering
    \includegraphics[width=\textwidth]{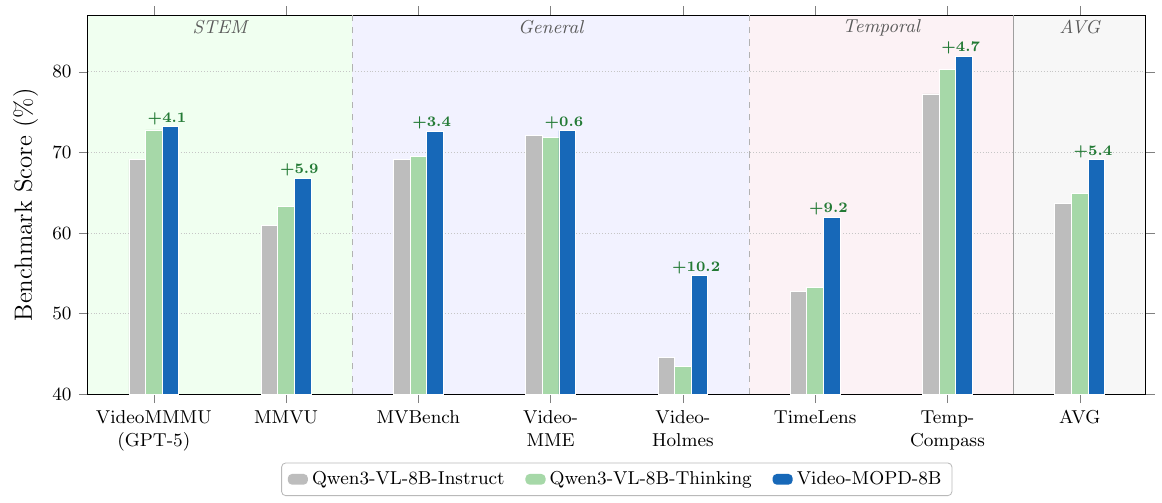}
    \caption{Overall comparison across three expert-aligned capability groups: STEM reasoning, general video understanding, and temporal grounding. Green annotations show the absolute gains of \model over Qwen3-VL-8B-Instruct.}
    \label{fig:model_comparison}
\end{figure}

\subsection{Main Results}

Figure~\ref{fig:model_comparison} shows that \model improves all seven benchmarks, with particularly large gains on Video-Holmes (+10.2), TimeLens (+9.2), and MMVU (+5.9), and raises the overall average by 5.4 points over Qwen3-VL-8B-Instruct. These results support the central claim of Video-MOPD: complementary expert policies can be combined into \model with gains that extend beyond any single training domain.

\subsection{Comprehensive Model Comparison}

\begin{table}[!t]
\centering
\caption{Comprehensive comparison across seven video benchmarks. VideoMMMU reports GPT-5 judge accuracy, and MMVU reports the GPT-5-judged overall score. Average is the arithmetic mean of the seven reported benchmark scores. Bold indicates the best result in each column, and underlining indicates the second-best result.}
\label{tab:comprehensive_results}
\scriptsize
\setlength{\tabcolsep}{1pt}
\renewcommand{\arraystretch}{1.08}
\begin{tabular*}{\textwidth}{@{\extracolsep{\fill}}lcccccccc@{}}
\toprule
\textbf{Model} & \textbf{VideoMMMU} & \textbf{MMVU} & \textbf{MVBench} & \textbf{TimeLens} & \textbf{Video-MME} & \textbf{TempCompass} & \textbf{Video-Holmes} & \textbf{Avg.} \\
\midrule
Qwen3-VL-8B-Instruct & 69.11 & 60.90 & 69.12 & 52.78 & 72.10 & 77.25 & 44.53 & 63.68 \\
Qwen3-VL-8B-Thinking & 72.78 & 63.30 & 69.45 & 53.29 & 71.90 & 80.24 & 43.49 & 64.92 \\
CRPO & 71.00 & 60.90 & 69.62 & 52.28 & 72.10 & \underline{82.40} & 46.54 & 64.98 \\
VideoKR & 64.44 & 62.00 & 66.50 & 27.66 & 68.50 & 79.90 & 47.36 & 59.48 \\
OneThinker & 65.89 & 61.80 & 69.50 & 43.03 & 67.00 & 80.21 & 50.35 & 62.54 \\
VideoSSR & 58.11 & 60.60 & 69.58 & 51.25 & 72.20 & 77.41 & 47.96 & 62.44 \\
\midrule
General Video Expert & 70.44 & 64.30 & \textbf{72.78} & 48.31 & \underline{72.60} & 81.10 & \underline{53.89} & 66.20 \\
STEM Expert & \textbf{73.33} & \underline{65.20} & 70.35 & 53.00 & 71.70 & \textbf{82.72} & 48.72 & 66.43 \\
VTG Expert & 58.89 & 60.40 & 67.33 & \textbf{62.32} & 68.80 & 77.21 & 45.89 & 62.98 \\
\midrule
Param-Merge (Avg.) & 71.44 & 64.30 & 70.45 & 60.46 & \textbf{72.70} & 81.98 & 49.65 & \underline{67.28} \\
\modelbf & \underline{73.22} & \textbf{66.80} & \underline{72.55} & \underline{61.98} & \textbf{72.70} & 81.91 & \textbf{54.71} & \textbf{69.12} \\
\bottomrule
\end{tabular*}
\end{table}

As Table~\ref{tab:comprehensive_results} shows, \model achieves the highest overall average of 69.12. Among non-expert baselines, it leads or ties on six of the seven benchmarks, including the best results on MMVU and Video-Holmes and a tie for the best Video-MME score. Compared with Param-Merge (Avg.), \model improves the overall average from 67.28 to 69.12, demonstrating the advantage of policy-space consolidation over direct parameter averaging. The three teachers exhibit complementary specialization---the General Video Expert excels at general video understanding, the STEM Expert at knowledge-intensive reasoning, and the VTG Expert at temporal localization---while \model integrates these strengths into a single deployable model with the strongest overall capability profile.

The cross-domain behavior further clarifies the complementary roles of the three teachers. The STEM Expert performs strongly on VideoMMMU and TempCompass, supporting our motivation that video reasoning frequently depends on fine-grained interpretation of individual frames. The VTG Expert contributes a highly specialized temporal-localization capability that complements the broader semantic and reasoning strengths of the General Video and STEM Experts. This complementary specialization motivates multi-teacher consolidation rather than relying on any single expert.

\subsection{Distillation Support Ablation}

\begin{table}[H]
\centering
\caption{Token-support ablation under one-epoch distillation. TimeLens uses a fixed 64-frame input. Bold marks the best score in each column.}
\label{tab:support_ablation}
\resizebox{\textwidth}{!}{%
\begin{tabular}{lcccccc}
\toprule
\textbf{Support} & \textbf{MVBench} & \textbf{MMVU} & \textbf{Video-MME} & \textbf{VideoMMMU} & \textbf{Video-Holmes} & \textbf{TimeLens} \\
\midrule
Sampled token & 72.45 & 66.50 & 72.8 & \textbf{73.67} & \textbf{50.57} & \textbf{61.42} \\
Teacher Top-16 & 72.20 & \textbf{68.00} & 73.0 & 71.44 & 50.41 & 60.92 \\
Student Top-16 & 72.32 & 66.70 & \textbf{73.4} & 72.11 & 50.35 & 60.43 \\
Teacher$\cap$Student Top-16 & \textbf{72.88} & 66.70 & 72.7 & 71.89 & 50.30 & 60.69 \\
\bottomrule
\end{tabular}}
\end{table}

As shown in Table~\ref{tab:support_ablation}, different token-support choices emphasize different regions of the teacher and student distributions, producing distinct transfer patterns across the evaluated capabilities. Video-MOPD therefore adopts sampled-token supervision as its default support.

\begin{figure}[H]
    \centering
    \includegraphics[width=\textwidth]{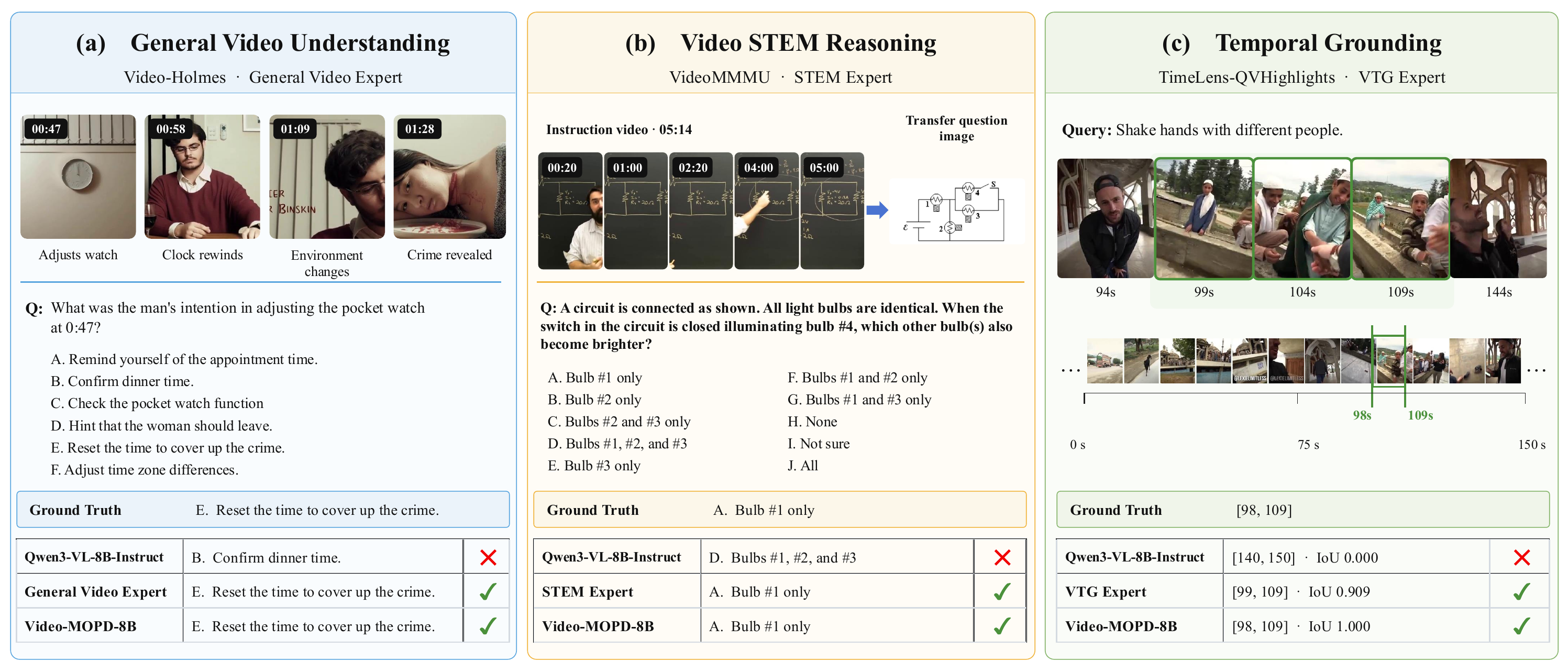}
    \caption{Qualitative comparison across general video understanding, video STEM reasoning, and temporal grounding. Each example compares Qwen3-VL-8B-Instruct, the matched expert, and \model.}
    \label{fig:qualitative_results}
\end{figure}

\subsection{Qualitative Analysis}
\label{sec:qualitative_analysis}

Figure~\ref{fig:qualitative_results} presents representative cases from the three expert-aligned capability domains: general video understanding, video STEM reasoning, and temporal grounding. Each case compares \model with the common base model and the corresponding domain expert. The examples show that \model reproduces expert-level behaviors across all three capability domains within a single model, covering cross-frame evidence integration, knowledge-intensive STEM reasoning, and precise temporal localization. In the temporal-grounding example, \model exactly recovers the ground-truth interval.

\section{Conclusion}
\label{section:conclusion}

We presented Video-MOPD, a specialize-then-unify framework for consolidating complementary video capabilities into \model. Starting from a shared base model, we construct specialized experts for general video understanding, frame-level STEM reasoning, and temporal grounding. To improve capability transfer, we introduce Reliability-Aware Informative Sampling (RAIS), which filters for reliable teacher supervision and prioritizes samples with larger teacher--student performance gaps. We then employ routed Multi-Teacher On-Policy Distillation to consolidate the complementary capabilities of these experts into a single unified model that requires neither teachers nor routing during inference. Across seven video benchmarks, \model improves the average score by 5.4 points over Qwen3-VL-8B-Instruct and reaches the highest overall score of 69.12. It further outperforms direct parameter averaging by 1.84 points, demonstrating the effectiveness of consolidating complementary specialists through reliability-aware, multi-teacher on-policy distillation. Video-MOPD therefore provides a unified post-training recipe for combining general video understanding, STEM reasoning, and precise temporal grounding in a single deployable model.

\section{Future Work}
\label{section:future_work}

Future work will extend Video-MOPD to long-form videos, where relevant evidence is sparse and distributed across distant temporal segments. We will explore efficient frame selection, hierarchical temporal representations, and memory-augmented reasoning for long-context processing. We also plan to strengthen the RL experts using broader training data, more accurate task-specific verifiers and rewards, and improved optimization algorithms, and to investigate whether iterative expert upgrading followed by repeated MOPD can transfer new capabilities without forgetting existing ones.

% \vspace{10mm}
\bibliography{neurips_2025}
\bibliographystyle{unsrtnat}

\clearpage
\appendix

\section{Detailed Training Configuration}
\label{app:training_details}

\begin{table}[H]
\centering
\caption{Main MOPD configuration.}
\label{tab:mopd_training_config}
\resizebox{0.88\textwidth}{!}{%
\begin{tabular}{lc}
\toprule
\textbf{Configuration} & \textbf{Value} \\
\midrule
Student initialization & Qwen3-VL-8B-Instruct \\
Mixture size & 7,493 \\
Video / Image / Temporal samples & 2,500 / 2,500 / 2,493 \\
Trainable modules & Language model \\
Frozen modules & Visual encoder and aligner \\
Learning rate & $5\times10^{-6}$ \\
Global batch size & 84 \\
Main training length & 2 epochs / 178 updates \\
Student generations per prompt & 1 \\
Generation temperature / top-$p$ & 1.0 / 1.0 \\
Maximum prompt / completion length & 18,432 / 4,096 \\
Total hardware & 4 nodes $\times$ 8 A800 GPUs \\
\bottomrule
\end{tabular}}
\end{table}

\section{Data Composition and Routing}
\label{app:data_routing}

\begin{table}[H]
\centering
\caption{Composition of the final MOPD training mixture.}
\label{tab:data_composition}
\resizebox{\textwidth}{!}{%
\begin{tabular}{lcll}
\toprule
\textbf{Domain} & \textbf{Count} & \textbf{Response type} & \textbf{Logical route} \\
\midrule
Video & 2,500 & Multiple choice with \texttt{<think>/<answer>} & General Video Expert \\
Image & 2,500 & Math and multiple-choice reasoning & STEM Expert \\
Temporal & 2,493 & Event start--end interval & VTG Expert \\
\bottomrule
\end{tabular}}
\end{table}

Routing uses the known data-domain label and is therefore deterministic. It should not be confused with a learned inference router. The label chooses the teacher service only during training, and it does not alter the student architecture.

\section{Detailed Evaluation Configuration}
\label{app:evaluation_configuration}

\paragraph{General video benchmarks.}
For all benchmarks except TimeLens, we use the vLLM inference backend and sample each video at 2 FPS, with at most 2,048 frames per example. Table~\ref{tab:general_evaluation_config} reports the shared decoding and inference configuration used for MVBench, MMVU, Video-MME, VideoMMMU, Video-Holmes, and TempCompass.

\begin{table}[H]
\centering
\caption{Evaluation configuration for all benchmarks except TimeLens.}
\label{tab:general_evaluation_config}
\begin{tabular}{lc}
\toprule
\textbf{Configuration} & \textbf{Value} \\
\midrule
Inference backend & vLLM \\
Video sampling rate & 2 FPS \\
\texttt{max\_frames} & 2,048 \\
\texttt{temperature} & 0.6 \\
\texttt{top\_p} & 0.8 \\
\texttt{top\_k} & 20 \\
\texttt{max\_tokens} & 32,768 \\
\texttt{repetition\_penalty} & 1.0 \\
\texttt{presence\_penalty} & 1.5 \\
Seed & 3,407 \\
\texttt{max\_num\_seqs} & 1 \\
\bottomrule
\end{tabular}
\end{table}

\paragraph{TimeLens.}
For TimeLens, we follow its dedicated temporal-localization evaluation pipeline using the Transformers backend. Videos are sampled at 4 FPS with a maximum of 2,048 frames. We use the official visual budget of $128{,}000 \times 32^2$ pixels and adopt greedy decoding with a maximum of 4,096 generated tokens. The key evaluation configuration is summarized in Table~\ref{tab:timelens_evaluation_config}.

\begin{table}[H]
\centering
\caption{Key evaluation configuration for TimeLens.}
\label{tab:timelens_evaluation_config}
\begin{tabular}{lc}
\toprule
\textbf{Configuration} & \textbf{Value} \\
\midrule
Inference backend & Transformers \\
Video sampling rate & 4 FPS \\
Maximum frames & 2,048 \\
Total visual budget & $128{,}000 \times 32^2$ pixels \\
Maximum generation length & 4,096 tokens \\
Decoding & Greedy ($\text{top-}k=1$) \\
\bottomrule
\end{tabular}
\end{table}

\section{Distillation-Support Variants}
\label{app:support_variants}

The sampled-token policy-gradient estimator of
Equation~\ref{eq:video_mopd} computes the teacher--student
log-probability difference for the token sampled by the student at
each position in the rollout. The Top-$k$ variants extend this
single-token support to candidate sets defined by the teacher,
the student, or their overlap. Teacher Top-$k$ uses the teacher's
highest-probability tokens, while Student Top-$k$ uses the student's
highest-probability tokens. Teacher$\cap$Student Top-$k$ restricts
the support to tokens appearing in both Top-$k$ sets. These variants
are evaluated under the same one-epoch setting in
Table~\ref{tab:support_ablation}.

\end{document}